%% file: acl2023.tex
\pdfoutput=1

\documentclass[11pt]{article}

\usepackage[final]{ACL2023}
\usepackage{hyperref}
\usepackage{times}
\usepackage{latexsym}
\usepackage{hyperref}
\usepackage{float}
\usepackage{subcaption}
\usepackage{xcolor}
\usepackage{float}

\usepackage[T1]{fontenc}

\usepackage[utf8]{inputenc}

\usepackage{microtype}
\usepackage{multirow}
\usepackage{listings}
\usepackage{inconsolata}
\newcommand{\methodname}[1]{GNOME}

\usepackage{adjustbox}
\usepackage{tabularx}
\title{GNOME: Generating Negotiations through Open-Domain Mapping of Exchanges}

\author{Darshan Deshpande$^{*, 1}$ \quad
Shambhavi Sinha$^{*, 1}$ \quad Anirudh Ravi Kumar$^{*, 1}$ \\ \textbf{Debaditya Pal}$^{*, 1}$ \quad \textbf{Jonathan May}$^{1,2}$ \\ 
${^1}$Thomas Lord Department of Computer Science, University of Southern California\\
${^2}$ Information Sciences Institute, University of Southern California\\
  \texttt{\{darshang, sinhasha, ar16847, debadity\}@usc.edu, jonmay@isi.edu}}

\begin{document}

\maketitle
\def\thefootnote{*}\footnotetext{These authors contributed equally to this work}\def\thefootnote{\arabic{footnote}}
\begin{abstract}
Language Models have previously shown strong negotiation capabilities in closed domains where the negotiation strategy prediction scope is constrained to a specific setup. In this paper, we first show that these models are not generalizable beyond their original training domain despite their wide-scale pretraining. Following this, we propose an automated framework called \methodname{}, which processes existing human-annotated, closed-domain datasets using Large Language Models and produces synthetic open-domain dialogues for negotiation. \methodname{} improves the generalizability of negotiation systems while reducing the expensive and subjective task of manual data curation. Through our experimental setup, we create a benchmark comparing encoder and decoder models trained on existing datasets against datasets created through \methodname{}. Our results show that models trained on our dataset not only perform better than previous state of the art models on domain specific strategy prediction, but also generalize better to previously unseen domains.
\end{abstract}

\section{Introduction}
\input{sections/introduction}
\section{GNOME as a Framework}
\input{sections/methodology}
\section{GNOME Dataset}
\input{sections/dataset}
\section{Experimental Setup}
\input{sections/experiments}

\section{Results}
\input{sections/results.tex}
\section{Relevant Work}
\input{sections/relevant_works}
\section{Conclusion}
\input{sections/conclusion}
\section*{Limitations}
\input{sections/limitations}
\section*{Ethics Statement}
\input{sections/ethics}

\bibliography{acl2023}
\bibliographystyle{acl_natbib}

\appendix

\section{Dataset Label Mappings}
\label{app:mappings}

\subsection{Casino Dataset}
\begin{lstlisting}
"Small-Talk":"Rapport",
"Empathy":"Rapport", 
"Coordination":"Coordination", 
"No-Need":"Self-Interest", 
"Elicit-Pref":"Coordination", 
"Undervalue- Partner": "Assessment", 
"Vouch-Fairness": "Assessment", 
"Other-Need":"Self-Interest", 
"Non-strategic": "Non-strategic"
\end{lstlisting}

\subsection{Craigslist Bargain Dataset}
\begin{lstlisting}
"intro":"Rapport",
propose": "Coordination", 
"vague-price":"Coordination", 
"counter":: "Coordination", 
"inform": "Assessment"
\end{lstlisting}

\subsection{Persuasion for Good Dataset}
\begin{lstlisting}
"Negotiate-Price-NoChange":"Coordination",
"Ask_Clarification-Y":"Self-Interest",
"Provide_Clarification-Y":"Self-Interest",
"tell_price":"Coordination",
"Negotiate-Remove-delivery":"Coordination",
"Ask_Price":"Coordination",
"Negotiate-Price-Decrease":"Coordination",
"Negotiate-Price-Increase":"Coordination",
"Acknowledge acceptance":"Assessment",
"Accept":"Assessment",
"Negotiate-Remove-X":"Coordination",
"Negotiate-Remove-X_Negotiate-Price-Decrease":"Coordination",
"Negotiate-Price-Remove-X":"Coordination",
"Negotiate-Add-X":"Coordination",
"Reject":"Assessment",
"Greet-Inform":"Rapport",
"Greet-Ask":"Rapport",
"Greet-Ask_Negotiate-Price-Decrease":"Rapport,Self-Interest",
"Greet-Inform_Negotiate-Price-Increase":"Rapport,Self-Interest",
"Greet-Inform_Negotiate-Price-NoChange":"Rapport,Self-Interest",
"Greet-Inform_Negotiate-Price-NoChange":"Rapport,Self-Interest",
"avoid_rejection":"Coordination"
\end{lstlisting}

\subsection{Job Interview Dataset}
\begin{lstlisting}
"greet":"Rapport",
"disagree":"Asessment",
"agree":"Asessment",
"inquire":"Coordination",
"propose":"Coordination",
"inform":"Asessment",
\end{lstlisting}

\section{Domain Adaptation Prompt}
\label{sec:appendix-prompt}
The following prompt was used to change the domain of the negotiation dialogs: \\

\noindent\textit{System: You are an assistant that can produce accurate and contextually grounded dialogues. You are tasked with changing the domain of the input dialogue.
In doing so, ONLY output the new domain in the format of NEW\_DOMAIN\{\{name\_of\_new\_domain\}\} followed by the domain-changed dialogue.}
\newline\newline
\noindent\textit{Prompt: Change the domain for this negotiation setting. Make sure to be as diverse but contextually grounded as possible when picking the new domain
and add a [EOS] token at the end of every utterance. Make sure to have same number of utterances in the output as there were in the given dialogue and DO NOT output any text other than the mapped dialogue: \{dialogue\}}

\section{Annotator Recruitment}
\label{app: Annotator Recruitement}
We recruited four annotators for our human evaluation studies from Amazon MTurk, a popular crowd-sourcing platform. The eligibility criteria were:

\begin{itemize}
    \item The individual must be over 18 years of age.
    \item The individual must be fluent in English.
\end{itemize}

For each survey completion, the participant was paid \$0.02. The requisite conditions are that they complete the survey. The payment and disbursement are handled internally by Amazon Mechanical Turk. The maximum allowed amount spent from discretionary on the annotation of all data points is \$10.

\newpage
\section{Skew Correction}
\begin{figure}[htbp]
    \centering
    \begin{subfigure}{\linewidth}
        \centering
        \includegraphics[width=\linewidth]{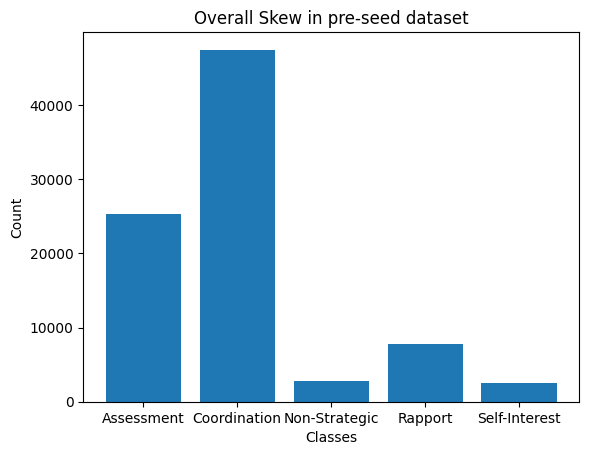}
        \caption{Observed skew before seed dataset creation}
        \label{fig:pre_seed_skew}
    \end{subfigure}
    \vspace{1em} 
    \begin{subfigure}{\linewidth}
        \centering
        \includegraphics[width=\linewidth]{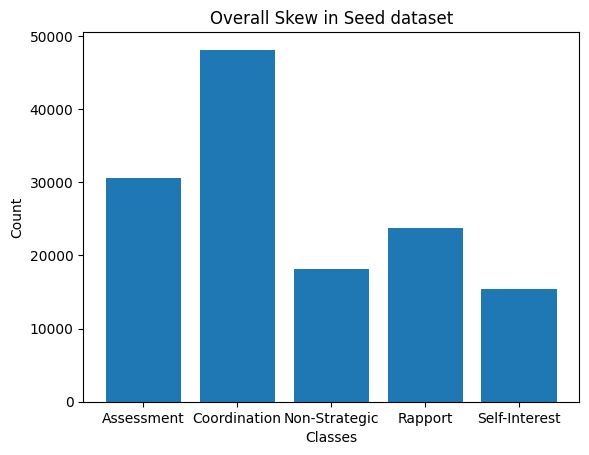}
        \caption{Observed skew after seed dataset creation}
        \label{fig:seed_skew}
    \end{subfigure}
\end{figure}
\section{Results}
\label{Resul}
    
    
    
    
\begin{figure*}[t]
    \centering
    \includegraphics[width=\textwidth]{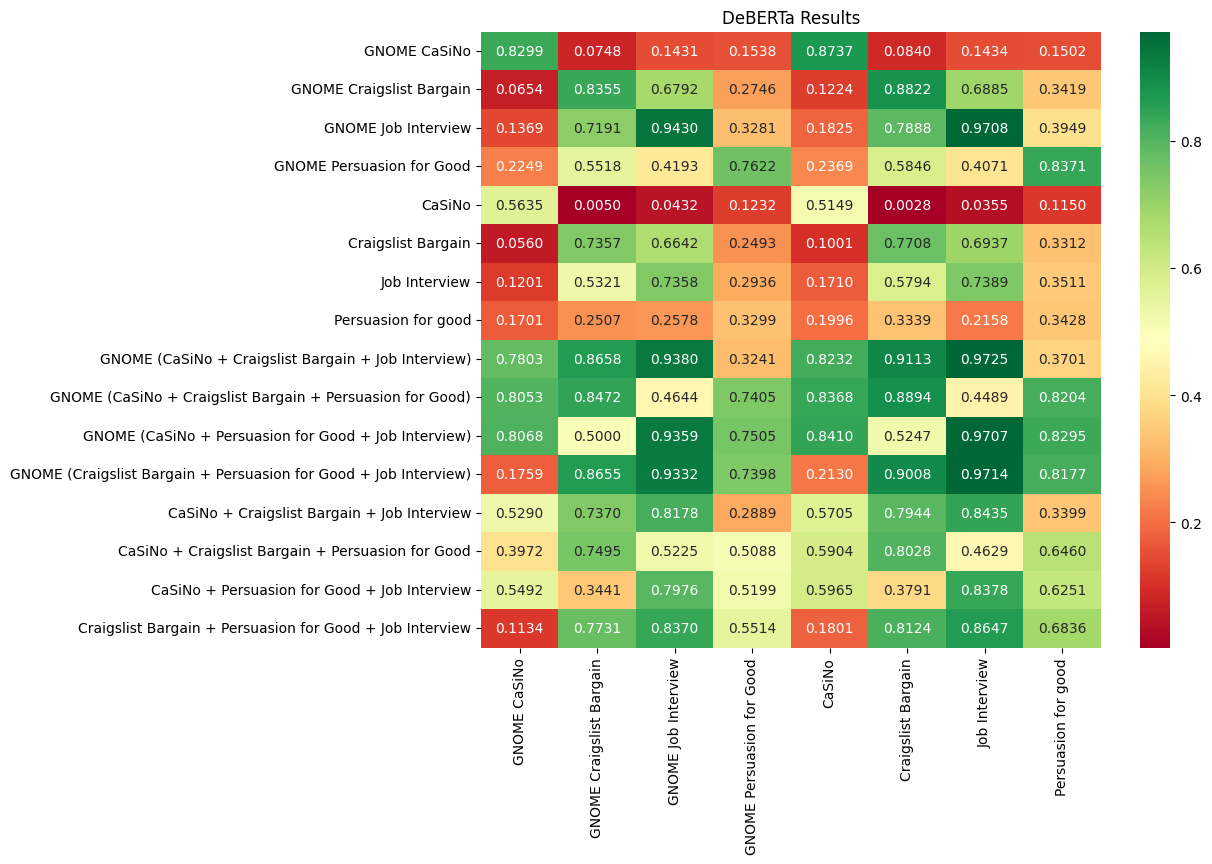}
    \caption{DeBERTa Results}
    \label{DeBERTa Results}
\end{figure*}
\vspace{-0.8em}
\begin{figure*}[t]
    \centering
    \includegraphics[width=\textwidth]{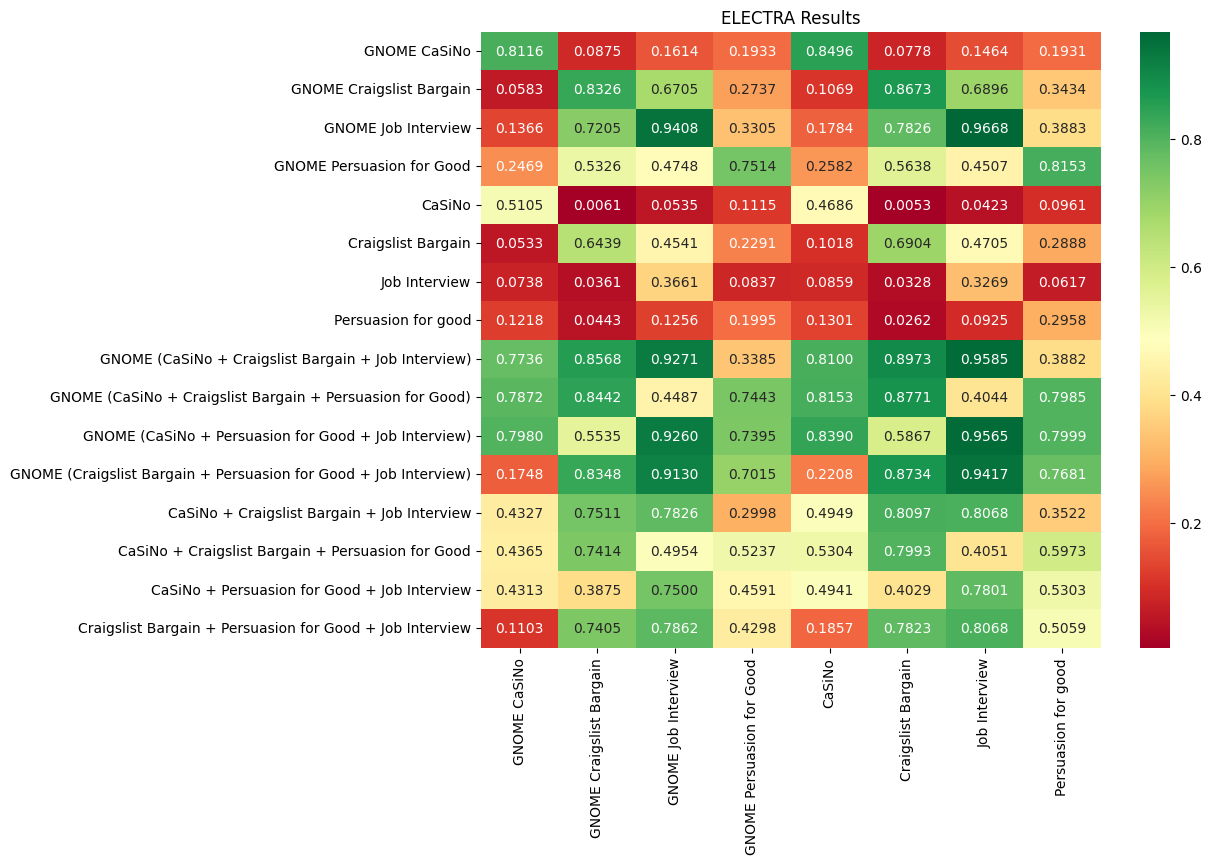}
    \caption{ELECTRA Results}
    \label{ELECTRA Results}
\end{figure*}

\begin{figure*}[t]
    \centering
    \includegraphics[width=\textwidth]{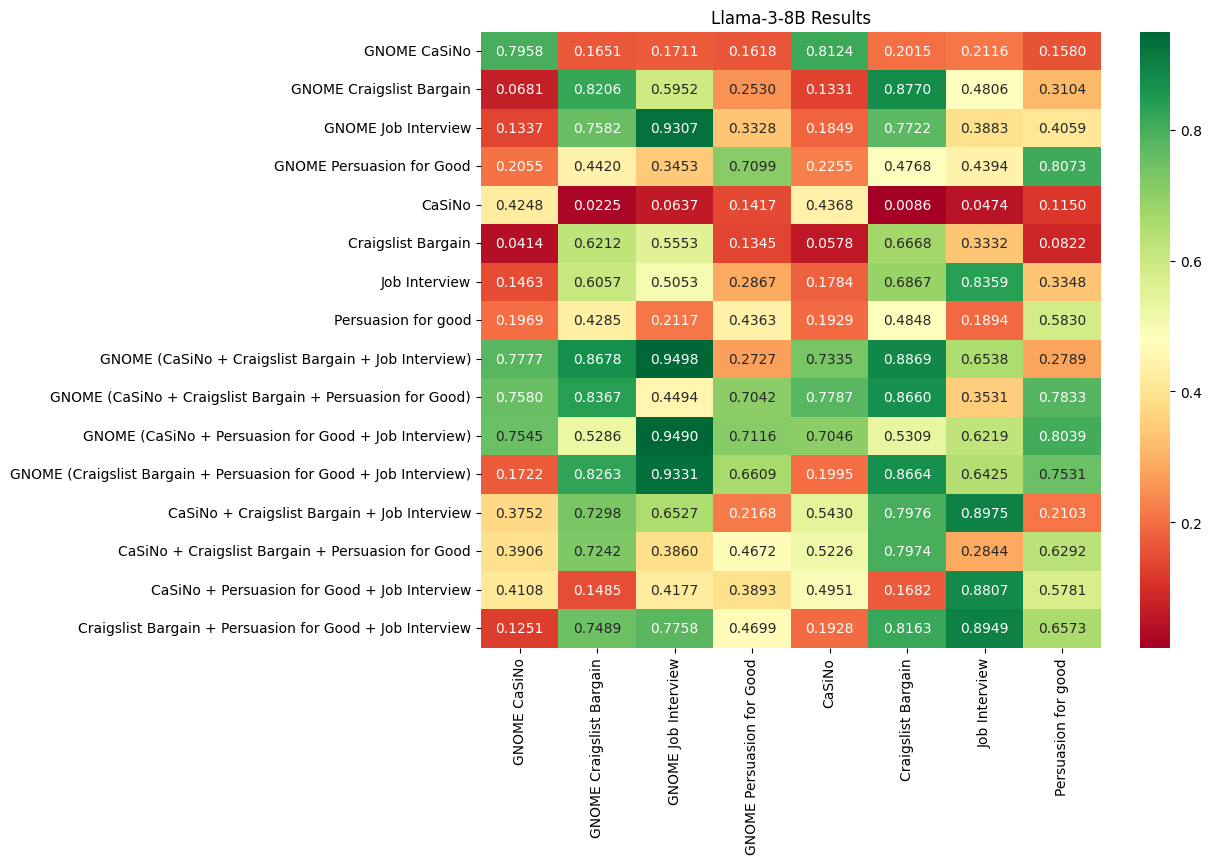}
    \caption{Llama-3-8B Results}
    \label{Llama-3-8B Results}
\end{figure*}

\begin{figure*}[t]
    \centering
    \includegraphics[width=\textwidth]{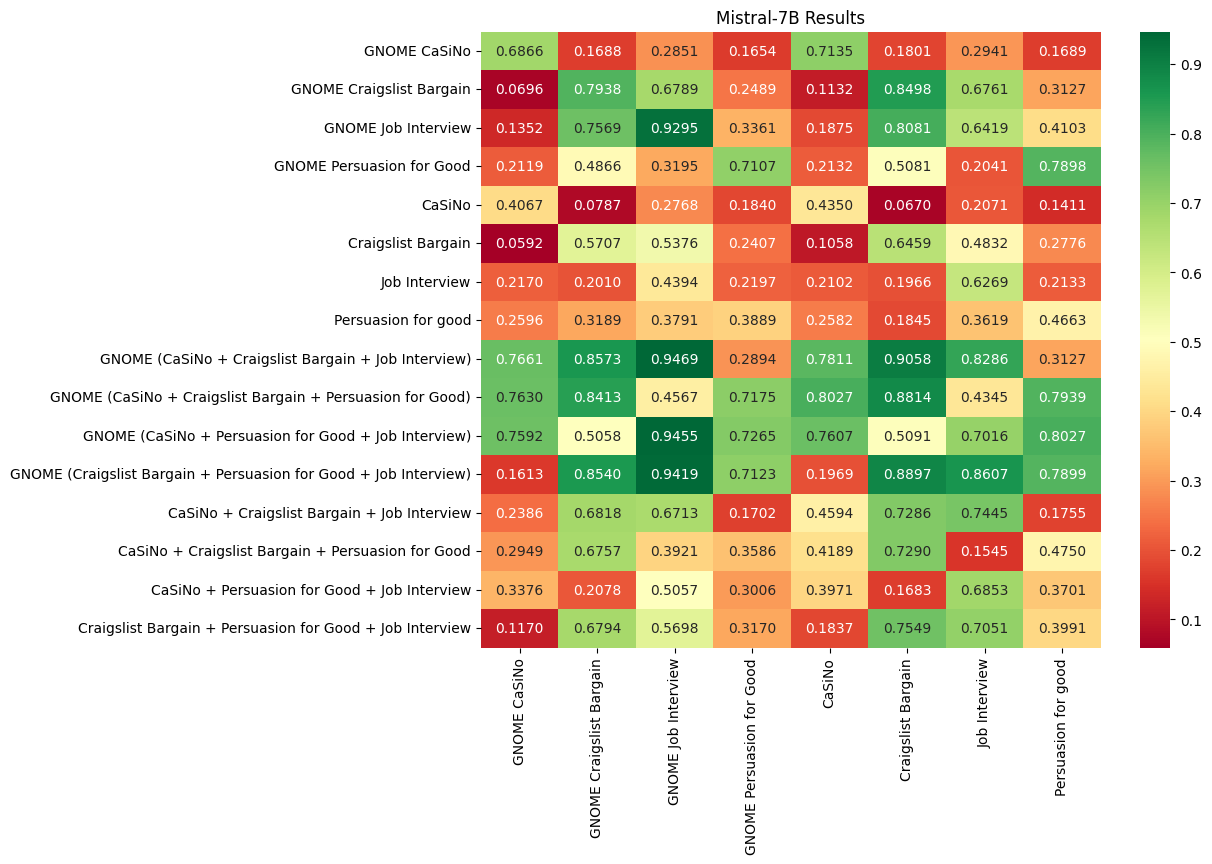}
    \caption{Mistral-7B Results}
    \label{Mistral-7B Results}
\end{figure*}

\end{document}

%% file: sections/introduction.tex

Negotiation is a key component of modern dialogue systems and has significant applications from bargaining~\cite{lewicki1981bargaining-application} to gaming~\cite{hausken1997game-application}, and business transactions~\cite{,  filzmoser2010automatedvshuman, mumpower1991judgment}. Negotiation agents powered by language models (LMs) are increasingly being used to conduct automated negotiations in a variety of contexts, including game theory~\cite{ding-catan-2021, peskov2020takes}, multi-issue bargaining~\cite{chawla2021casino, job_interview} and pyschotherapy~\cite{tanana2016comparison}.
\begin{figure}
    \includegraphics[width=0.5\textwidth]{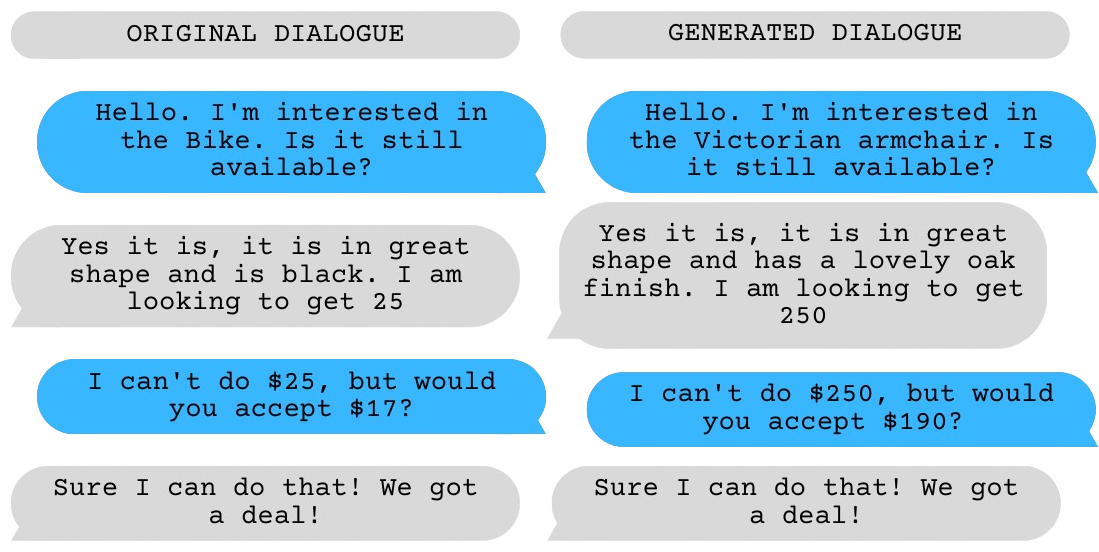}
    \caption{GNOME generated negotiation dialogue}
    \label{fig:overview}
    \vspace{-1em}
\end{figure}

\begin{figure*}
    \centering
    \includegraphics[width=\linewidth]{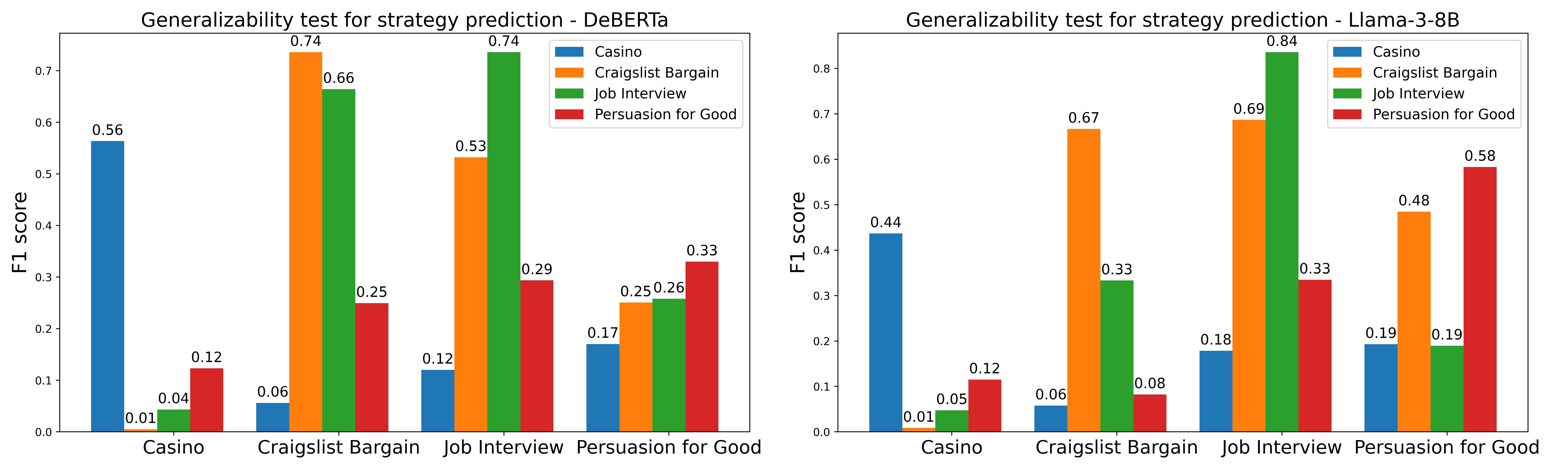}
    \caption{Generalizability test for existing datasets}
     \label{fig:generalizability_results}
     \vspace{-0.5em}
\end{figure*}

Previous works have explored models trained for negotiation strategy prediction that aim to label each utterance in a dialogue with the appropriate negotiation strategies to enable the negotiator to understand the opponent's motive effectively.
These studies include
development of strategy prediction models for tasks such as commodity price bargaining~\cite{ahmad2023ina, craiglist_bargain} and item negotiation in a barter setting~\cite{lewis2017dealnodeal, chawla2021casino} while limiting the scope of the negotiation to a specific selection of items (closed-domain). However, to facilitate the wider adoption and domain-agnostic negotiation understanding, the training datasets must be as varied and open as possible~\cite{craiglist_bargain}.
Additionally, closed-domain settings have been shown to affect out-of-domain task performance, which remains an unsolved challenge for modern negotiation agents~\cite{li-murray-2023-zero}.

Recently, Large Language Models (LLMs) have shown excellent generalization capabilities in zero and few-shot settings across multiple benchmarks~\cite{bubeck2023sparks, chang2023survey,kojima2022large, hou2024large, ahmed2022few, dai2022promptagator}. As a preliminary experiment to our study, we show that this generalization capability is impacted when LLMs are fine-tuned on closed-domain negotiation strategy prediction tasks. For our experiments, we selected four popular datasets, trained models on these individually, and evaluated their performance on out-of-domain data from other datasets. These datasets included CaSiNo~\cite{chawla2021casino}, which consists of negotiations involving campfire resources like firewood, water, and food; Job Interview~\cite{job_interview}, containing simulated job interview negotiations for salary and responsibilities; Craigslist Bargain~\cite{craiglist_bargain}, focused on negotiations for items on the Craigslist platform; and Persuasion For Good~\cite{persuasion-for-good}, where participants are persuaded to donate to social causes. These datasets present a varied set of negotiation strategies that were manually mapped to a more general set of labels to allow for cross-comparison (more details about this mapping can be found in \autoref{app:mappings}). 

Our results presented in \autoref{fig:generalizability_results} demonstrate that existing encoders and LLMs trained on individual datasets show poor generalization when tested on datasets with different negotiation scenarios. Furthermore, we observed that domain-restricted training, for example, on item exchange-based CaSiNo, poorly generalizes on monetary negotiations, as found in the Job Interview and Craigslist Bargain datasets. This highlights the need for diverse, cross-domain training data.

To address this generalizability problem, we introduce a novel framework for \textbf{G}enerating \textbf{N}egotiations through \textbf{O}pen \textbf{D}omain \textbf{M}apping of \textbf{E}xchanges, or \methodname{}, that utilizes existing human-annotated closed-domain datasets to produce synthetic open-domain datasets for strategy prediction for negotiation. An example of this is shown in \autoref{fig:overview} where the original dialogue for a bike sale is translated to a negotiation for a Victorian armchair while maintaining the consistency of the original dialogue.
\begin{figure*}
    \centering
    \includegraphics[width=\textwidth]{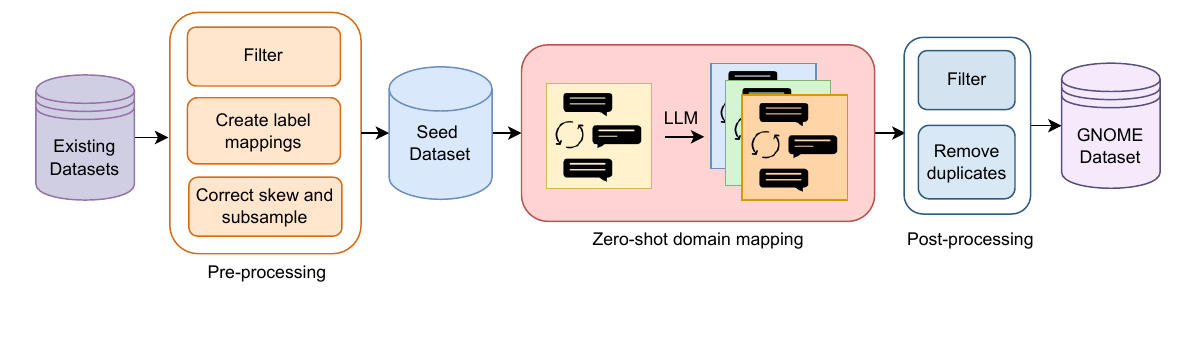}
    \vspace{-3.5em}
    \caption{GNOME Framework}
    \label{fig:gnome_framework}
\end{figure*}
In this work, we show that:
\begin{enumerate}
    \item Models trained and fine-tuned on open-domain data exhibit improved performance in domain-specific strategy prediction tasks.
    \item Open-domain training and fine-tuning enhances a model's ability to generalize to novel, unseen domains.
    \item High-quality synthetic data, derived from human annotations, can effectively substitute for costly additional human-annotated data.
\end{enumerate} 

%% file: sections/methodology.tex
\label{sec: framework}
The GNOME framework has three distinct stages: pre-processing, domain mapping and post-processing, as illustrated in \autoref{fig:gnome_framework}. This section describes the framework and methodology employed for the curation of the GNOME generated dataset or the GNOME dataset.

\paragraph{Pre-processing:}
The pre-processing stage consists of three sub-stages. 

First, we filter out incomplete dialogues as we believe they do not contain enough strategic information for accurate strategy prediction.

The second sub-stage involves creating common label mappings for strategies across datasets in order to evaluate GNOME's out-of-domain performance and generalizability. In order to do so, we group similar labels according to their definition. The labels for these groups then become the new labels. The label mapping ensures that our analysis is both inclusive of the specific nuances within the datasets and broadly applicable to different negotiation settings. A few instances of the common mappings are provided in \autoref{table:mappings}. Here, we present examples of semantically similar utterances that demonstrate coordination among negotiators. Despite having different labels in their respective datasets, these utterances are mapped to a common \textit{Coordination} strategy during our label-mapping stage. A detailed breakdown of the mapped label instances can be found in  \autoref{app:mappings}. 
\begin{table*}[ht!]
\centering
\centering
\begin{adjustbox}{max width=\linewidth}
\begin{tabular}{ m{4cm}  m{7cm}  m{3cm}  m{3cm} }
\textbf{Dataset} & \textbf{Utterance} & \textbf{Original Label} & \textbf{Mapped Label} \\[1mm]
\hline
Craigslist Bargain & The wheels seem nice on it, but they could be a beter quality. Would you accept 100 for the board? & propose & Coordination \\
\hline
CaSiNo & We need some firewood too, though! Let's try to make a deal that benefits us both! Could I have 1 firewood, 3 food, and 3 waters? & Coordination & Coordination \\
\hline
Job Interview & you have something for 5 days per week any job with above average salary ? & inquire & Coordination \\
\hline
Persuasion for Good & Thank you.Can we make it \$1.50?These children really need the assistance. & Negotiate-Price-Increase & Coordination \\
\hline
\end{tabular}
\end{adjustbox}
\caption{Example of Label Mapping}
\label{table:mappings}
\end{table*}

The final sub-stage involves skew correction of the dataset where we aim to counter the imbalance in the distribution that results from the combination of the datasets. This is necessary to ensure that equal importance is assigned to all negotiation strategies. 
Each dialogue in the dataset consists of multiple turns and has multiple labels associated with each turn. Since we need to sample entire dialogues, we must consider the label distribution of entire dialogue as opposed to individual utterances. To achieve this, we employ a frequency-based sampling approach where each label is assigned a score inversely proportional to its frequency within the dataset. The score for each utterance is then computed as the sum of the individual label scores. Finally, the score for each dialogue instance is derived by summing the scores of its constituent utterances. After scoring the datasets, we select the top $k$ highest-scoring dialogue instances from each dataset and combine them to form the seed dataset for our experiments.

\paragraph{Domain Mapping:}
Domain mapping refers to changing the negotiation setting of dialogues to a more diverse and unique scenario. This adaptation is achieved by feeding individual dialogues to a Large Language Model in a zero-shot manner. We map the entire dialogue with a random seed and a temperature of one at every generation step to maximize diversity and discourage repetition. 
For completeness, we experimented with generating utterances individually by feeding the dialogue till the current turn. However, this approach resulted in decreased generation quality and consistency across utterances. We attribute this inconsistency to the restricted context window accessible to LLM. Hence, we map the entire dialogue at once.
\vspace{-0.1em}\\
\indent The prompt used for this task is defined in \autoref{sec:appendix-prompt}. Our proposed prompt incorporates several constraints to promote diversity and contextual coherence in the target domain. Firstly, we ensure that the generated dialogue adheres to the style established in the original dialogue and maintains fluency. 
To preserve the labels between the original and domain-mapped utterance, we assert that the length of the generated dialogue must equal the length of the original dialogue.
Further, we instruct the model to append an [EOS] token to each generated utterance, making it convenient for automated parsing and label transfer to the newly generated dialogue. Finally, as an optional step to ease the qualitative analysis, we prompt the model to generate the topic of the new domain. We repeat the above process $n$ times over the seed dataset to produce approximately the same number of dialogues as all original datasets combined. 

\paragraph{Post-processing:} To refine the generated data, we remove the dialogues shorter than the original instances, which indicate the LLM's failure to generate the designated [EOS] token.
Furthermore, we identify and remove any potential duplicate dialogues produced during generation. We also ensure that the dataset does not contain instances from the seed dataset to prevent data leakage.

%% file: sections/dataset.tex
\label{sec: dataset}
In this section, we describe the curation parameters and characteristics of the open-domain synthetic dataset we created using \methodname{}. We select four manually annotated negotiation datasets: the CaSiNo corpus \cite{chawla2021casino}, the Craigslist Bargain Dataset \cite{craiglist_bargain}, the Job Interview Dataset \cite{job_interview} and the Persuasion for Good corpus \cite{persuasion-for-good}. These datasets were chosen as they cover a wide range of negotiation settings like multi-item bargaining, price negotiation, integrative negotiation and persuasion. The number of dialogues in each dataset is presented in table \ref{tab: num dialogue}.

\begin{table}[!ht]
\centering
\begin{adjustbox}{width=0.8\linewidth}
\begin{tabular}{lr}
\textbf{Dataset}    & \textbf{\# of Dialogues} \\ \hline
CaSiNo              & 396                      \\ 
Craigslist Bargain  & 6555                     \\ 
Job Interview       & 2577                     \\ 
Persuasion for Good & 300                      \\ 
\hline
\end{tabular}
\end{adjustbox}
\caption{Number of dialogues in each dataset before seed dataset creation}
\label{tab: num dialogue}
\end{table}

After selecting the datasets, the next step is to map them to a common set of labels so that the same strategy prediction model can be trained and evaluated on all of them. The process of label mapping is carried out manually in order to ensure good quality labels that covers all of the different strategies in these datasets. After manual review, we find that the various labels can be mapped to a common set of five labels, which are: Rapport, Assessment, Self-Interest, Coordination and Non-Strategic. These labels were chosen because they accurately cover the diverse range of strategy labels present in the datasets used and are comprehensive enough to potentially represent the strategies used in any negotiation setting. After the label mapping procedure, we observe a significant skew in the combined dataset prior to forming the seed dataset (Figure \ref{fig:pre_seed_skew}). Hence, we use the skew correction method described in \autoref{fig:gnome_framework} in an effort to balance the label distributions. We then sample $k=250$ dialogues from each of the four datasets to curate the seed dataset. The observed skew in the seed dataset is reported in \autoref{fig:seed_skew}.

For the domain mapping stage, we choose Llama-3-70B~\cite{llama3modelcard} as it is the leading open-source LLM in its size category \cite{llama3modelcard}. We then map the seed dataset $n=10$ times to generate approximately the same number of dialogues as all original datasets combined (\autoref{tab: num dialogue}). The resulting dataset contains 288 duplicates, which are removed in the post-processing step. We also observe inconsistencies in instruction following, resulting in misplaced [EOS] tokens in the generated dialogue. This misplacement causes errors in dialogue extraction and label assignment during post-processing, and hence, we remove these instances. We report the total number of misplaced tokens in \autoref{tab: misplaced eos}.

\begin{table}[!ht]
\begin{adjustbox}{width=\linewidth}
\begin{tabular}{lr}
\textbf{Dataset}    & \textbf{\# of misplaced {[}EOS{]} tokens} \\ \hline
CaSiNo              & 236                                       \\ 
Craigslist Bargain  & 247                                       \\ 
Job Interview       & 30                                        \\ 
Persuasion for Good & 227                                       \\ \hline
\end{tabular}
\end{adjustbox}
\caption{Number of misplaced [EOS] tokens after domain-mapping}
\label{tab: misplaced eos}
\end{table}

In order to discover the overall number of domains we utilized clustering analysis on the SentenceTransformer embeddings \cite{reimers-2019-sentence-bert} of the generated domain titles. Through the analysis, we observed the presence of 472 distinct domains, which were additionally verified through manual inspection for overlap. 

\subsection{Human Evaluations}

The human evaluation study analyzes the outputs of the domain mapping stage. We take a random sample (without replacement) of 100 dialogue pairs, consisting of the original dialogue and the domain-mapped dialogue, with uniform probability and use it for the study. The dialogue pairs consist of the original dialogue as well as the domain mapped dialogue. We crowdsourced four annotators for this study (\autoref{app: Annotator Recruitement}) who were shown the pairs one at a time. The annotators are asked to rate the dialogue pair in terms of structural similarity and coherence. Structural similarity refers to how similar the domain-mapped dialogue is compared to the original dialogue. This is an important evaluation metric as we plan to utilize the same labels from the original dialogue for the domain-mapped one. Coherence is defined as the overall understandability, congruity and consistency of the domain mapped dialogue. The annotators are asked to rate both of these metrics on a Likert scale ranging from 1 to 5. After the annotation process, we calculate Krippendorff's alpha to measure inter-annotator reliability. We further present the average metric values and percentage agreement from the annotations to indicate dataset quality in~\autoref{tab:dom_mapping_reliability}.

%% file: sections/experiments.tex
In this section, we present our experimental setup for evaluating the impact of \methodname{} on the generalization capabilities of various models for the strategy prediction task. We focus on the DeBERTa-v3\footnote{\url{https://huggingface.co/microsoft/deberta-v3-base}} and ELECTRA\footnote{\url{https://huggingface.co/google/electra-base-discriminator}} for encoder models, and Llama-3-8B\footnote{\url{https://huggingface.co/meta-llama/Meta-Llama-3-8B}} and Mistral\footnote{\url{https://huggingface.co/mistralai/Mistral-7B-v0.1}}\cite{jiang2023mistral} for decoder LLM models as they are strong baselines for their respective categories \cite{he2023debertav3, DBLP:journals/corr/abs-2003-10555,llama3modelcard, jiang2023mistral}. We fine-tune our models on a single NVIDIA A100 GPU using a weighted cross-entropy loss to address data imbalance in the multi-label text classification task.

For DeBERTa and ELECTRA, the fine-tuning process involves 3 epochs, a batch size of 16, and a cosine learning rate scheduler with a learning rate of 8e-4 for DeBERTa and a learning rate of 4e-4 for ELECTRA. For Llama-3-8B and Mistral, we apply 4-bit quantization using bitsandbytes \cite{dettmers2022optimizers} to reduce the model size. Additionally, we use LoRA (Low-Rank Adaptation) fine-tuning \cite{hu2021lora} with Flash Attention 2 \cite{dao2023flashattention2} for faster training. The LoRA fine-tuning is performed with a rank and alpha of 16 over 3 epochs. We set the batch size to 16, the gradient accumulation steps to 8 and the learning rate to 5e-4 for Llama and 6e-4 for Mistral using the last token for classification. The hyperparameters for these models were exhaustively searched to optimize performance.

We contrast the performance of a model fine-tuned on \methodname{}-augmented data (\textit{\methodname{} A}) against the original data \textit{A} when tested on \textit{A} to demonstrate that training in a general setting improves task-specific performance. We call these experiments synthetic-in-domain and in-domain experiments respectively. We then evaluate if a model trained on \textit{\methodname{} A} can generalize well to an unseen dataset (\textit{B}), which we refer to as synthetic-out-of-domain test. Additionally, we evaluate the generalization performance of the models by training the models on combined datasets \textit{(A + B + C)} and testing them on an unseen dataset (D) as a leave-one-out experiment. Here \textit{(A, B, C, D)} are the datasets mentioned in \autoref{sec: dataset}. All the datasets have a 60-40 train-validation split. For evaluation, we calculated weighted F1 scores and class-wise joint accuracy to measure performance on the imbalanced dataset. Class-wise joint accuracy measures the overall correctness of predictions per class in a multi-label classification setting.

%% file: sections/results.tex
\begin{table*}[!ht]
\centering
\begin{adjustbox}{max width=\linewidth}
\begin{tabular}{m{4cm} m{5cm} rrrr}
    \textbf{Experiment} & \textbf{Fine-tune $\rightarrow$ Test} & \textbf{DeBERTa} &
    \textbf{ELECTRA} &
    \textbf{Llama-3-8B} &
    \textbf{Mistral-7B}\\
    \hline
    In-domain & A $\rightarrow$ A & 0.5918 & 0.4454 & 0.6306 & 0.5435 \\
    Synthetic-in-domain & \methodname{} A $\rightarrow$ A & \textbf{0.8910} & \textbf{0.8747} & \textbf{0.7213} & \textbf{0.7487}\\
    \hline
    Out-of-domain & A $\rightarrow$ B & 0.2539 & 0.1169 & 0.2223 & 0.2244 \\
    Synthetic-out-of-domain & \methodname{} A $\rightarrow$ B & \textbf{0.3438} & \textbf{0.3483} & \textbf{0.3333} & \textbf{0.3397}\\
    \hline
    Leave-one-out & A + B + C $\rightarrow$ D & 0.3405 & 0.3365 & 0.2139 & 0.1705\\
    Synthetic-leave-one-out & \methodname{} (A + B + C) $\rightarrow$ D & \textbf{0.3892} & \textbf{0.4000} & \textbf{0.3406} & \textbf{0.3633}\\
   \hline
\end{tabular}
\end{adjustbox}
\caption{Average F1 scores reported on corresponding experiments. A,B,C,D $\in$ \{CaSiNo, Craigslist Bargain, Job Interview, Persuasion for Good\}. The individiual F1 scores are taken from the respective experiments and then averaged. \protect\footnotemark}

\label{tab:results}
\end{table*}

We evaluate the effectiveness of the \methodname{} framework in addressing our research questions: \textit{Do models fine-tuned on open-domain data show improved performance in domain-specific strategy prediction? (RQ1)}, \textit{Does open-domain training and fine-tuning improve a model's generalizability? (RQ2)} and \textit{Can \methodname{} data substitute human-annotated data? (RQ3) }

\paragraph{Do models fine-tuned on open-domain data show improved performance in domain-specific strategy prediction? (RQ1)}
The first block in \autoref{tab:results} represents the results of in-domain ad synthetic-in-domain experiments. We observe that DeBERTa, ELECTRA, Llama-3-8B, and Mistral-7B exhibit absolute increase in F1 scores of 29.91\%,  42.93\% , 9.07\%, and 20.52\%, respectively, when trained on \methodname{} datasets compared to their performance when trained on their original versions. Due to the wide range of domains covered by \methodname{} data, the models are exposed to more 
nuanced representations and various linguistic patterns, which make them perform better on domain-specific test sets. Hence, we conclude that the model trained on  \textit{\methodname{} A} significantly enhances task-specific performance compared to fine-tuning on the corresponding original dataset (\textit{A}). Additionally, we perform a Qualitative analysis to confirm that the improvements in F1 score is due to added diversity in \autoref{sec: QA}.

\paragraph{Does open-domain training and finetuning improve a model's generalizability? (RQ2)}

The second block in \autoref{tab:results} shows the results from the out-of-domain and synthetic-out-of-domain experiments. We observe that DeBERTa, ELECTRA, Llama-3-8B, and Mistral-7B exhibit absolute increase in F1 scores of 8.9\%,  23.14\% , 12.67\%, and 11.53\%, respectively, when trained on \methodname{} datasets compared to their performance when trained on original datasets. An interesting observation is that Mistral-7B displays better generalization than Llama-3-8B on both the original and \methodname{}-augmented datasets despite Llama-3-8B's larger pre-training. This suggests that the size of the pre-training corpus is independent of downstream performance on strategy prediction. 

Furthermore, the third block in \autoref{tab:results} represents the results of the leave-one-out and synthetic-leave-one-out experiments. We see that DeBERTa, ELECTRA, Llama-3-8B, and Mistral-7B exhibit absolute increase in F1 scores of  4.87\%,  6.35\% , 11.11\%, and 19.28\%, respectively, when trained on \methodname{} datasets compared to their performance when trained on a combination of original datasets. We also observe that encoder models generalize better than the decoder models when trained on \textit{\methodname{} (A + B + C)}. 

Concluding from the above two experiments, we can see that the models fine-tuned on GNOME datasets generalize better to an unseen domain than the ones trained on domain-specific datasets.

\footnotetext{All reported results are statistically significant across 3 runs. For detailed results, refer to \autoref{Resul}.}

\begin{table*}[!ht]
    \centering
    \small
    \begin{tabular}{p{0.47\textwidth} p{0.10\textwidth} p{0.14\textwidth} p{0.16\textwidth}}
\textbf{Utterance} & \textbf{True \newline Label} & \textbf{GNOME Model \newline Label} & \textbf{Original Dataset Model Label}  \\
\hline 
HI! Are you excited to launch our new fashion line?\newline I am \newline ...\newline \emph{What sort of rare fabrics are you looking for?}                                                                & Coordination             & Coordination                                                               & Self-Interest                                                                        \\ \hline
Hello\newline Do you have a preference for any extra amenities for your flight?\newline Yes. I need 2 extra legroom seats, 2 lounge passes, and 1 meal upgrade\newline ...\newline \emph{Thank you for your time} & Rapport                  & Assessment                                                                 & Rapport                                                                              \\ \hline
Hello\newline I was wondering if I can rent the luxury sedan and the SUV\newline ...\newline \emph{I am an avid traveler. i can afford to give you the luxury sedan for 2 days for the SUV for 3 days though} & Coordination             & Coordination                                                              & Coordination                                                                        \\ \hline
\end{tabular}
\caption{Qualitative Analysis of model performance on GNOME dataset. The utterance for which the label is defined is italicized.}
\label{tab: qualitative}
\end{table*}

\paragraph{Can \methodname{} data substitute human-annotated data for negotiations? (RQ3)}

\begin{table}[!ht]
\begin{adjustbox}{width=\linewidth}
\begin{tabular}{lrr}
                    \textbf{Metric} & \textbf{Structural Similarity} & \textbf{Coherence}         \\ \hline
Krippendorf's alpha &   0.6819                       & 0.5557                     \\ 
Average Value  &     4.808     & 4.615 \\ 
Percentage Agreement  &     85\%     & 55\% \\ 
\hline
\end{tabular}
\end{adjustbox}
\caption{The table shows the annotator reliability metrics for the GNOME dataset. Average values calculated on a scale of 1-5.}
\label{tab:dom_mapping_reliability}
\vspace{-1.3em}
\end{table}

\autoref{tab:dom_mapping_reliability} describes metrics from the human evaluation studies performed on the \methodname{} dataset. We observe that the average coherence and structural similarity annotation values are greater than 4.5 out of 5, indicating that the generated responses are logical, congruent, and grounded in real-life negotiation dialogues in structure. We noticed some instances with high structural similarity but low coherence scores, which indicates that the domain mapping step was able to follow the original dialogue structurally, but the original dialogue was of poor quality. We found the Job Interview dataset to contain the majority of such incoherent dialogues while randomly sampling for human evaluations.

\citet{VANDERLEE2021101151} present a survey of observed annotator reliability metrics for automatically generated texts where the authors find that most of the reported alpha values fall in the range of 0.3 - 0.5 for such approaches. \citet{li2023synthetic} also report similar trends and show that the alpha values decrease with an increase in the subjectivity of the datasets. Our calculated alpha values are higher than this range, which indicates a better agreement compared to other synthetic datasets. To further assess the level of agreement, we calculate the percentage agreement of the Likert scale annotations. The coherence and structural similarity metrics have a percentage agreement of 55\% and 85\%, respectively, which indicates that the annotators have high agreement considering the incoherence introduced by the Job Interview dataset. Considering the three reported values in \autoref{tab:dom_mapping_reliability}, we believe that we have a strong inter-annotator agreement. Furthermore, based on the high average annotation values, we can claim that \methodname{} data can effectively substitute human-annotated data.

Referring to our research questions, the results discussed above, confirm that \methodname{} datasets help improve performance on domain-specific negotiations while promoting generalizability across domains. Finally, we show that \methodname{} data is an effective substitute for human-annotated data.

\section{Qualitative Analysis}
\label{sec: QA}
\autoref{tab: qualitative} displays example utterances from the \methodname{} dataset, the true labels, and the predicted labels from a model fine-tuned on the \methodname{} dataset and a model trained on the original dataset. The first row contains a dialogue that was mapped from the CaSiNo dataset. During domain mapping, the items being negotiated were changed from the scope of the CaSiNo dataset to rare fabrics and other fashion accessories. The model trained on the original CaSiNo dataset was not exposed to any examples involving such items and, therefore, makes a mistake in classifying the utterance. The model trained on the \methodname{} dataset was introduced to a variety of negotiation scenarios, and it can generalize better to previously unseen items. It can, thus, correctly classify this utterance. 

The second example highlights a unique condition where the label of the utterance is not dependent on the context of the negotiation. Utterances like "Hello" and "Thank you" are examples of rapport building, irrespective of how the rest of the dialogue continues. The model trained on the original dataset accurately identifies this condition and correctly classifies this utterance. The model trained on the \methodname{} dataset is exposed to a greater variety of negotiation scenarios, and a significant portion of them contain assessment utterances that start with a polite refusal of a previously made offer. For example, "Thank you, but I cannot accept this deal" leads to confusion and incorrect classification.

The final example shows how the structure of the utterance can help both models provide accurate results. Usually, during negotiations, numbers only come up while the participants are trying to coordinate a bargain. For example: "I can pay \$20 for this tshirt". In such examples, the model can simply recognize multiple numbers as a pattern to accurately classify them as "Coordination" utterances.

%% file: sections/relevant_works.tex
\paragraph{Negotiation Datasets} 
The ability to engage in effective negotiations is an important skill for intelligent agents, enabling them to reach agreeable outcomes through persuasive dialogue across various contexts~\cite{lets-negotiate-2024}. Through the domain of in-game negotiation settings, \citet{lewis2017dealnodeal} were among the first to explore end-to-end learning of negotiation dialogues. They trained models directly from human-human negotiations without annotated dialogue states, creating the Deal Or No Deal dataset with 5,808 dialogues. Following this, in an effort to decouple strategy prediction and negotiation generation \citet{craiglist_bargain} released the Craigslist Bargain dataset, which includes multiple buyer-seller negotiation scenarios from Craigslist Bargain and human-annotated negotiation strategy labels, expanding the scope of negotiation research to include strategic elements and grounding it in real-world concepts. Extending the work on datasets grounded in real-world concepts, \citet{persuasion-for-good} introduced Persuasion for Good, a dataset aimed at developing personalized persuasive dialogue systems for social good, such as charitable donations. This dataset highlights the potential of negotiation dialogue systems to contribute to socially beneficial outcomes.

In an effort to shift the focus of negotiation datasets from single-issue to multi-issue bargaining \citet{chawla2021casino} introduced CaSiNo, a corpus of 1,030 negotiation dialogues set in a camping scenario and \citet{job_interview} released a corpus of job interviews dealing with negotiations regarding salary and responsibilities. More recently, \citet{INA-dataset-2023} developed an Integrative Negotiation Agent (INA) by prompting GPT-J. This agent is designed for integrative negotiations, dynamically adjusting prices and negotiating the inclusion or exclusion of items in a bundle deal, showcasing advanced negotiation capabilities.

The datasets discussed above focused solely on bilateral dialogue. To introduce a different negotiation setting, \citet{ding-catan-2021} introduced DinG, a French corpus capturing multi-party negotiations during Settlers of Catan gameplay.

\paragraph{Synthetic Datasets} 
Previous work involving the use of LLM-generated synthetic data includes \citet{li2023synthetic}, where the authors attempted to benchmark text classification models trained on synthetic data. However, they found that using synthetic data lead to instability in model performance depending on the subjectivity of the classification problem. Higher subjectivity at both the task and instance level was negatively correlated with model performance. \citet{liu2024best} and \citet{bao-etal-2023-synthetic} describe synthetic data generation pipelines to generate high-quality domain-specific data. The authors further show that high-quality synthetic data grounded in human-annotated datasets can significantly improve model performance.

With \methodname{}, we combine different types of negotiation (e.g., multi-issue bargaining, persuasion, price negotiation) from multiple datasets while creating diverse negotiation scenarios through domain mapping. As observed by \citet{shu2021open}, open-domain training can improve model generalizability on previously unseen datasets. Furthermore, by grounding the synthetic data in human-annotated datasets, we ensure high-quality generations. 

\paragraph{Large Language Models (LLMs) for Negotiation}
With the recent emergence of Large Language Models (LLMs) like Meta's Llama \cite{llama3modelcard} and OpenAI's ChatGPT \cite{OpenAI_GPT4_2024}, there have been several studies that experiment with the negotiation capabilities of such models \cite{zhan2022lets}. \citet{fu2023improving} uses LLMs in a self-play setting where the models negotiate with and improve each other in a game setting. The authors find that only a subset of LLMs can improve in a self-play setting with minimal human intervention. \citet{xu2024exploring} analyze the performance of LLMs in the context of the negotiation based communication game "Werewolf" and show that the models are effectively able to play the game without the need for fine-tuning. \citet{chen2024money} created an evaluation suite to test the negotiation capabilities of LLMs in an auction setting and reported that the models possess key skills for auction negotiation which can further be improved using adaptive strategies. However, the authors find that occasionally the LLMs would be outperformed by simpler methods indicating opportunities for further advancement.

%% file: sections/conclusion.tex
\vspace{-0.3em}
In this work, we introduce \methodname{}, a framework designed to generate synthetic open-domain negotiation dialogues from existing human-annotated closed-domain datasets. \methodname{} demonstrates that it is possible to generate diverse and domain-independent datasets, which in turn enhances the performance and generalization abilities of negotiation models. This approach addresses the limitations of current negotiation models, which often fail to generalize beyond the specific contexts of their training data. Our benchmarking compares encoder and decoder models trained on traditional datasets versus those generated by \methodname{}, revealing significant improvements in the ability of models to generalize to new negotiation scenarios. This demonstrates the efficacy of synthetic data in reducing reliance on extensive human annotation, which is often subjective and labor-intensive. In the future, we aim to focus on expanding the variety of negotiation scenarios and further refining the synthetic data generation process to enhance the realism and applicability of the dialogues produced.
Additionally, we encourage research on replacing \methodname{}'s dependency on manual mapping of labels with an automated labeling procedure. This extension will further validate the flexibility and utility of \methodname{} in creating high-quality synthetic data across a wide range of applications. We also hope to extend \methodname{} beyond negotiations, exploring its potential applications in other dialogue-driven domains and tasks.

%% file: sections/limitations.tex
Although GNOME aims to produce diverse and high-quality synthetic dialogues, the quality of generated data is inherently dependent on the original closed-domain datasets and the large language models (LLMs) used. Any biases or limitations present in the source datasets can propagate into the synthetic data, potentially affecting the performance and generalizability of the trained models. Even though GNOME automates the generation of negotiation dialogues, the quality of synthetic data can vary. The generated dialogues might not always capture the nuances and subtleties of human negotiations, potentially affecting the performance of models trained on this data.

%% file: sections/ethics.tex
We only use openly available datasets (CaSiNo: CC-BY-4.0, Job Interview, Craigslist Bargain: MIT License, Persuasion For Good: Apache 2.0) and encourage users of GNOME to strictly adhere to the licensing of the datasets used. We do not collect any personally identifiable data during the creation of the GNOME dataset. Instead, we relied on publicly accessible datasets that were previously annotated for negotiation scenarios.
In the qualitative evaluation of the data, all human annotations were obtained with appropriate consent and we ensured that the use of these annotations complied with ethical standards and data protection regulations. The above study was exempted from the IRB review (UP-24-00493).
Further, due to the automated nature of the generation process, there remains a possibility that offensive content could be inadvertently generated. We encourage users to utilize safety tools like Llama Guard ~\citep{llama-guard} to ensure safer generations.
By addressing these considerations, we aim to promote the ethical use of GNOME and contribute to the development of fair and responsible negotiation models.